\documentclass[preprint,12pt,authoryear]{elsarticle}
\usepackage{amssymb}
\usepackage{amsmath}
\usepackage{booktabs}
\usepackage{longtable}
\usepackage{graphicx}
\usepackage{float}
\usepackage{tabularx}

\journal{Journal of Biomedical Informatics}

\begin{document}
\newpageafter{author}

\begin{frontmatter}

\title{\textit{FairLogue}: A Toolkit for Intersectional Fairness Analysis in Clinical Machine Learning Models}

\author[1]{Nick Souligne, M.S.\corref{cor1}}
\ead{nasouligne@arizona.edu}

\author[1]{Vignesh Subbian, PhD.}

\affiliation[1]{organization={Department of Biomedical Engineering, University of Arizona},
  addressline={1127 E. James E. Rogers Way},
  city={Tucson},
  state={AZ},
  postcode={85721-0020},
  country={United States}}

\cortext[cor1]{Corresponding Author}

\begin{abstract}

\textbf{Objective:}
Algorithmic fairness is essential for ensuring equitable and trustworthy machine learning applications in healthcare. Existing fairness tools primarily focus on single-axis demographic comparisons and may fail to identify compounded disparities affecting intersectional populations. This study introduces \textit{Fairlogue}, a toolkit to operationalize intersectional fairness assessment in observational and counterfactual scenarios within real-world clinical contexts.

\textbf{Methods:}
\textit{Fairlogue} is implemented as a Python-based toolkit composed of three components: (1) an observational fairness framework extending demographic parity, equalized odds, and equal opportunity difference to intersectional scenarios; (2) a counterfactual and intersectional framework assessing fairness under treatment-based contexts; and (3) a generalized counterfactual framework assessing fairness under intersectional group membership interventions. The toolkit was evaluated using electronic health record data from the \textit{All of Us} Controlled Tier V8 dataset for a glaucoma surgery prediction task involving a logistic regression model with intersectional protected attributes race and gender.

\textbf{Results:}
Observational analysis identified substantial intersectional disparities despite moderate overall model performance (AUROC = 0.709; accuracy = 0.651). Intersectional evaluation revealed larger fairness gaps compared to single-axis analyses, including demographic parity differences of 0.20 and equalized odds true positive rate and false positive rate gaps of 0.33 and 0.15, respectively. Counterfactual analysis using permutation-based null distributions produced overall measures of unfairness (``u-values'') approaching zero, suggesting that conditional on covariates, observed disparities were consistent with chance rather than systematic dependence on group membership; however, interpretation required caution due to weak associations between protected characteristics and outcomes.

\textbf{Conclusion:}
\textit{Fairlogue} provides a practical, modular toolkit for quantifying intersectional biases in applied clinical machine learning workflows. In providing observational and counterfactual analyses, the toolkit enables comprehensive auditing of intersectional disparities while highlighting how single-axis fairness evaluations may underestimate disparities, and facilitates accessible, reproducible assessments across health and other domains involving sensitive attributes.

\end{abstract}

\end{frontmatter}
\clearpage


\section{Introduction}
Algorithmic fairness, the detection and handling of biases in machine learning models, is a critical component of developing trustworthy and reliable models. Predictive models hold considerable promise for improving patient outcomes and healthcare delivery, but without safeguards such as appropriate bias mitigation methods, they risk perpetuating and amplifying existing inequities in the training data, particularly for historically marginalized populations \citep{hort2023}. Importantly, these inequities often arise from the sociotechnical processes through which clinical data is generated. Prior work has highlighted how factors such as social stigma, inconsistent diagnostic labeling, and structural disparities in healthcare access shape the composition and quality of behavioral health and clinical datasets, introducing bias before model development begins \citep{walsh2020}. Furthermore, persistence of bias in the model training data without mitigation may limit the generalizability of the model to new datasets and contexts. For example, a glaucoma intervention prediction model trained using data from a single site with limited sociodemographic diversity failed to generalize to a larger, more heterogenous population due to unaddressed imbalance in the training data \citep{baxter2021}. Additionally, even when trained on the larger heterogenous dataset, the model performance varied significantly between demographic subgroups. Similar results have been found in a variety of clinical contexts and tasks including survival analysis and disease prediction \citep{rahman2022,poulain2023}. These findings highlight that algorithmic fairness techniques are prerequisite for creating reliable models deployable in clinical settings. \par

In response to these challenges, a growing number of methods have emerged to assess and mitigate biases throughout the development cycle. These methods broadly fall into three categories: pre-processing (e.g., reweighting or sampling adjustments), in-processing (e.g., fairness-constrained optimization), or post-processing (e.g., threshold adjustments). While these techniques often reduce certain forms of bias, they frequently rely on comparisons across single demographic attributes and may fail to detect or address the compounded disparities experienced by individuals belonging to multiple underrepresented groups. As research has shown, if a fairness measure fails to account for certain phenomenon (such as a mismatch between group representation in treatments and data availability between groups), then the measure will fail and could falsely indicate the model is fair, when it in fact discriminates against at least one group \citep{wastvedt2023}. This limitation highlights the need for new fairness approaches that account for intersectional biases. \par

Intersectionality based approaches recognize that the experiences and outcomes for each individual are shaped by overlapping social identities, and failure to account for these interactions may lead to misleading conclusions regarding model fairness. Emerging research suggests that intersectional bias quantification and mitigation strategies can not only improve model performance for individual subgroups but also maintain high overall predictive performances \citep{wastvedt2023, foulds2019, lacava2023}. These results indicate that intersecting sources of biases can persist despite commonly used fairness techniques and highlight the need for mitigation approaches that explicitly account for intersectionality. \par

Electronic health record (EHR) based biomedical datasets are promising resources for developing and evaluating machine learning models, offering large-scale clinical data that can support predictive modeling across diverse clinical contexts. Furthermore, the size and longitudinal nature of these datasets enable analysis of complex patient trajectories, treatment patterns, and outcomes. As predictive models trained on EHR data are frequently translated into clinical contexts, fairness auditing in these datasets directly informs safe and equitable real-world deployment. Unfortunately, not all of these datasets are focused on collating data from a diverse population which may limit the generalizability to new data of any model trained on them. With a focus on recruiting traditionally underrepresented populations and the scale of the dataset enabling analysis of intersectional subgroups that may otherwise be too sparsely represented for reliable estimation, the \textit{All of Us Research Program} is uniquely situated for algorithmic fairness investigations \citep{allofus}. Additionally, there is a substantial body of literature that developed risk prediction or classification models using the \textit{All of Us} data. Despite this wealth of research, very few studies have looked to assess algorithmic fairness in this context, and none have systematically assessed intersectional fairness in models using the dataset. Therefore, the goals of this study are (1) to develop a practical toolkit, called \textit{FairLogue}, that operationalizes quantification and contextualization of intersectional biases in real-world biomedical data and models and (2) to demonstrate the application of the toolkit to a real-world clinical case study using \textit{All of Us} data and illustrate how intersectional fairness assessments can reveal disparities beyond traditional single-axis evaluations.

\subsection*{Statement of significance}
\begin{table}[!ht]
\centering
\small
\setlength{\tabcolsep}{6pt}
\renewcommand{\arraystretch}{1.2}
\begin{tabular}{p{0.28\textwidth} p{0.66\textwidth}}
\toprule
\textbf{Problem or Issue} &
Traditional algorithmic fairness evaluations in clinical machine learning primarily rely on single-axis demographic comparisons, which may overlook compounded disparities affecting intersectional subgroups and limit equitable deployment of predictive models. \\
\midrule
\textbf{What is Already Known} &
Existing fairness toolkits and auditing methods can detect bias across individual protected attributes, and counterfactual frameworks have been proposed; however, practical implementations that integrate intersectional analysis with real-world clinical workflows remain limited. \\
\midrule
\textbf{What This Paper Adds} &
This paper introduces \textit{FairLogue}, a modular toolkit that combines observational and counterfactual intersectional fairness assessments within a unified framework. Applied to \textit{All of Us} clinical data, the toolkit demonstrates that intersectional analyses can reveal disparities not captured by single-axis evaluations and provides practical workflows for diagnosing whether disparities reflect model unfairness or underlying data structure. \\
\midrule
\textbf{Who would benefit from the new knowledge in this paper} &
Clinical informaticians, machine learning researchers, healthcare data scientists, and developers deploying predictive models who require practical methods to evaluate and improve fairness across diverse patient populations. \\
\bottomrule
\end{tabular}
\end{table}

\section{Related Work}
A growing ecosystem of analytical tools have been developed to support fairness evaluation and mitigation in machine learning, although their capabilities vary considerably in terms of supported metrics, mitigation approaches, and applicability to healthcare contexts. Widely used Python libraries such as AIF360 \citep{aif360}, Aequitas \citep{aequitas}, Fairkit-learn \citep{fairkitlearn}, and Fairlearn \citep{fairlearn} are exemplar packages that enable bias quantification and a limited number of bias mitigation strategies. Other notable Python libraries include Fairness Indicators (as part of the Tensorflow \citep{tensorflow} ecosystem) and Fairlens \citep{fairlens}, both of which focus on bias detection and quantification. In the R ecosystem, there are packages such as Fairmodels \citep{fairmodels} or Fairness \citep{fairnessR} that similarly allow for bias detection or mitigation strategies. \par

While these tools provide important infrastructure for identifying potential bias, they are largely designed around single-attribute group comparisons and do not natively support more complex assessments involving multiple demographic attributes simultaneously. This limitation is significant because fairness approaches that ignore intersectionality run the risk of obscuring disparities that affect marginalized subgroups. Several studies have shown that failure to account for this limitation may produce misleading conclusions that fail to account for differences in data availability, treatment representation, or subgroup representation for intersectional populations \citep{wastvedt2023, foulds2019, lacava2023}. As a result, existing tools often fall short of addressing the full breadth of algorithmic fairness challenges. \par

There have been several studies of note demonstrating both the need for intersectional fairness considerations and how to quantify these biases in clinical contexts. For instance, an auditing study assessed the fairness and reliability of multiple advance care planning models across three settings: primary care, inpatient oncology, and hospital medicine \citep{lu2022}. Upon comparing results with clinician-generated reports on patient mortality, significant disparities were found along intersectional demographic axes. Similarly, a systematic study of vision-language foundational models for predicting demographic attributes from chest x-ray images revealed pronounced intersectional biases that were not reflected in evaluations by board-certified radiologists \citep{yang2024}. There is also an extensive body of work around single-axis algorithmic fairness in various contexts, as described in a comprehensive review by \citet{hort2023}, which reinforce that current, single axis approaches are insufficient to fully address the scope of biases.\par

Among emerging algorithmic fairness approaches, a small number of studies have explicitly included intersectionality into the formulations. One such study introduces the concept of differential fairness, inspired by the idea of differential privacy, that defines fairness guarantees across all demographic intersections simultaneously \citep{foulds2019}. This in-processing regularization technique employs differential fairness as a penalty term to bound disparities between subgroup combinations via probabilistic constraints, in effect protecting small or overlapping demographic groups that are often overlooked by more traditional fairness metrics. While this approach holds promise, reported implementations utilize a single model type (neural network) and on limited clinical data, which limits the generalizability and reproducibility of the technique. \par

Another study of note is a fairness-aware framework that integrates algorithmic fairness with interpretable modeling techniques, emphasizing transparency and trustworthiness in healthcare decision support systems \citep{faim2024}. This framework operationalizes optimal predictive model selection under several different scenarios including full, partial, and no awareness of either single or multiple protected characteristics. Demonstrating how fairness considerations and interpretability can be incorporated to improve clinical applicability, this framework represents a bridge between these concepts and provides a path forward for future work. However, there are limitations in the application of intersectionality. As the framework is focused on model selection and not any sort of intervention on intersectional characteristics beyond omitting them from the potential models, it is likely that the optimal models will resort to “fairness through unawareness” as reported in the empirical results. It is generally held that fairness through unawareness is insufficient at best, and ineffective or even harmful at worst due to limitations such as latent or proxy variables \citep{barocas2023}. \par

In addition to the above-mentioned observational fairness methods, causal and counterfactual approaches have gained attention as a means of evaluating fairness beyond correlational comparisons. Notably, \citet{wastvedt2023} introduced a framework that conceptualizes fairness through counterfactual performance metrics and supports evaluation of disparities by estimating outcomes under alternative demographic conditions while holding relevant covariates constant. This approach represents an important advancement beyond traditional group fairness methods by incorporating causal reasoning into fairness assessment and enabling more nuanced evaluation of disparities across intersecting identities. Despite these strengths, practical adoption of this framework remains limited by several constraints. First, existing implementations are not readily integrated into standard machine learning workflows, requiring substantial adaptation for applied research settings. Second, the framework assumes the presence of explicit treatment variables or predefined causal structures, limiting applicability in scenarios where protected attributes themselves serve as the primary axis of evaluation. These limitations restrict the ability of researchers to operationalize counterfactual intersectional fairness analyses in real-world clinical settings \par

Taken together, the existing literature on intersectional approaches within clinical contexts represent a growing recognition of the necessity for incorporating intersectionality. These tools and frameworks are generally perceived to be useful, but with limitations, including the lack of practical implementations. Addressing this gap requires new frameworks that integrate both observational and counterfactual perspectives, while remaining practical and accessible for real-world research pipelines and clinical implementations. \par

The \textit{Fairlogue} toolkit builds upon these developments by operationalizing fairness analysis into three complementary components. Component 1 (intersectional fairness metrics) provides an observational assessment using both single-axis and intersectional extensions of widely used fairness metrics and enables users to identify disparities without imposing causal assumptions. Component 2 (counterfactual intersectional fairness) operationalizes prior research by translating theoretical formulations of counterfactual fairness frameworks into a modular analytical workflow for machine learning pipelines. Lastly, Component 3 further generalizes the counterfactual fairness framework to a wide range of models by relaxing assumptions regarding explicit treatment variables and extending counterfactual analysis to settings where intersectional characteristics may serve as the means of intervention. Together, these components aim to bridge the gap between theoretical advances in intersectional fairness and practical implementation within real-world healthcare settings. 


\section{Methods}
Each component in the \textit{Fairlogue} toolkit corresponds to a different methodological approach to quantifying intersectional biases. This design approach was chosen to emphasize modularity in allowing the user to pick and choose which components are appropriate for their use case. Within each component, there are separate modules detailing class definitions, auxiliary functions, and reference implementations. Additionally, each module contains markdown files with documentation including detailed explanations of each function, parameter descriptions, return types, and behavioral specifications. The core functionality and output of each component is summarized below. 

\subsection{Component 1: Intersectional Fairness Metrics}
The first component of the toolkit provides a generalized pre-processing framework for assessing common fairness metrics in both a single-axis and intersectional manner. The single-axis metrics are intended to serve as a baseline for comparison with the intersectional metrics. In extending the traditional single-axis fairness metrics, the framework allows the user to easily examine how models perform across combinations of protected characteristics (e.g., gender x race) in comparison to the model performance on single-axis metrics. These metrics include Demographic Parity (DP), Equalized Odds (EO), and Equal Opportunity Difference (EOD). Demographic Parity evaluates whether the probability of receiving a positive prediction is equivalent across all groups, Equalized Odds tests whether true positive rates (TPR) and false positive rates (FPR) are balanced across groups, and Equal Opportunity Difference focuses on disparities in TPR. These metrics were chosen because they are widely used in algorithmic fairness literature, but the toolkit does leave room for additional metrics to be added. Additional performance metrics such as area under the receiver operating characteristic curve (AUROC) and accuracy are also reported. \par

The workflow of the component starts with the user supplying a tabular dataset containing at least two protected attributes, outcome variable, and covariates. The user must then specify a model type (e.g., Logistic Regression, Random Forest, Decision Tree, LightGBM, or Neural Network), model hyperparameters, classification thresholds, and optional class imbalance controls. During execution, the framework constructs an intersectional group identifier by imputing the protected characteristics and (optionally) filters out underrepresented subgroups with representation under a provided threshold to prevent unstable estimates (e.g., subgroups with near-zero representation). The data is then split into training and testing data, trained on the supplied features, predicts on the test set, and finally computes the confusion matrix generating the true positives (TP), false positives (FP), true negatives (TN), and false negatives (FN). From the confusion matrix, the TPR, FPR, and prediction rates are derived for each intersectional group, then aggregated into the previously mentioned fairness metrics. The output of this framework contains a structured results table with each metric, broken down by demographic subgroup, and visualizations displaying group-level metrics to enable straightforward comparison of disparities between groups. \par

This component is intended as the most accessible entry point for users and the first step in quantifying intersectional biases in predictive models by providing a descriptive baseline before introducing counterfactual or causal predictions. User-specified parameters for enforcing class balance ensure that fairness metrics remain interpretable even when subgroup sample sizes are limited. Additional information on the individual functions and available user configurations are further described in the toolkit documentation. A general workflow description of this component can be found in Figure~\ref{component1} below. 

\begin{figure}[H]
\centering
\includegraphics[width=\linewidth, keepaspectratio]{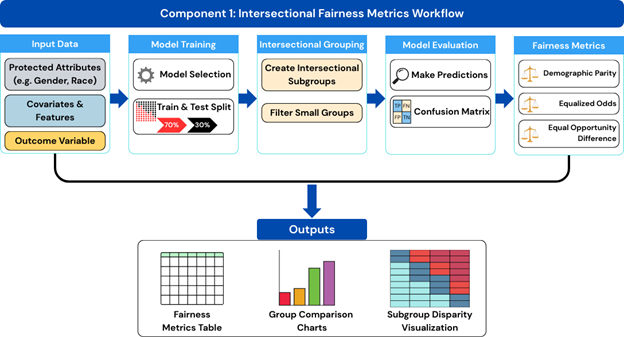}
\caption{\textbf{Component 1 workflow for observational intersectional fairness assessment}. Input data is used to train a predictive model and compute single-axis and intersectional fairness metrics (demographic parity, equalized odds, equal opportunity difference), producing group-level disparities summaries without causal assumptions.}
\label{component1}
\end{figure}

\subsection{Component 2: Counterfactual Intersectional Fairness (with Treatment)}
The second component implements a post-processing framework in which intersectional biases are quantified in clinical treatment-based models under a counterfactual group assignment. While Component 1 focuses on quantifying observational disparities in model predictions, Component 2 evaluates counterfactual disparities or how predictions differ if the individual were randomly assigned different intersectional identities while holding treatment status and covariates constant. The implementation presented here is a faithful translation from R to Python of the original framework introduced by \citet{wastvedt2023}. \par

The original framework estimates pairwise differences in model prediction error rates across intersectional groups under a counterfactual group membership assignment, with the treatment exposure as the causal variable for interest. The formulation for the error rates are given below in equation~\ref{eq:empirical_cf}. 

\begin{center}
\textbf{Variables:}

\vspace{0.5em}

\begin{tabular}{l l}
$A$   & = binary protected characteristic \\
$S$   & = binary risk prediction score \\
$D$   & = binary treatment assignment \\
$Y$   & = binary outcome such as adverse health event \\
$Y_0$ & = binary outcome under no treatment ($D = 0$) \\
$Y_1$ & = binary outcome under treatment ($D = 1$) \\
\end{tabular}
\end{center}

\noindent\textbf{Definition 1:} counterfactual false positive rate ($\mathbf{cFPR}(S,a)$) and false negative rate ($\mathbf{cFNR}(S,a)$) are:\\
\[
\mathbf{cFPR}(S,a) = \Pr\!\left(S = 1 \,\middle|\, Y_0 = 0,\ A = a\right), \
\mathbf{cFNR}(S,a) = \Pr\!\left(S = 0 \,\middle|\, Y_0 = 1,\ A = a\right)
\]
\noindent\textbf{Empirically, these are:}

\begin{equation}
\label{eq:empirical_cf}
\mathbf{cFPR}
=
\frac{\mathbb{E}\!\left[S(1-Y)(1-D)\right]}
     {\mathbb{E}\!\left[(1-Y)(1-D)\right]},
\qquad
\mathbf{cFNR}
=
\frac{\mathbb{E}\!\left[(1-S)Y_0\right]}
     {\mathbb{E}\!\left[Y_0\right]}
\end{equation}

The framework then compares the observed disparities with those expected under a null distribution representing a hypothetically perfectly fair model. The null distribution is generated by randomly permuting group membership to each individual, then estimating nuisance parameters and predicting the outcome variable. This has the effect of decoupling the effects of group membership from the outcome, thus simulating a hypothetically, perfectly fair world. Optional bootstrapping resampling procedures are available to help quantify uncertainty, and data borrowing options can be used to improve stability in small samples. \par

The framework generates a table of unfairness values, or “u-values”, where smaller values indicate the observed disparities are consistent with chance, and large values indicate systemic unfairness relative to the null benchmark. The user is also able to provide a threshold value, indicating the acceptable level of unfairness relative to the null distribution in order to balance the well-known problem of bias-accuracy trade-off. A high-level overview of the component is demonstrated below in Figure~\ref{counterfactual}. The user first fits a model of the outcome under the observed treatment, then passes the predicted probabilities and metadata to the framework. The component constructs the null distributions (via random group membership permutations), optionally performs bootstrap resampling, estimates per-group error rates, and finally outputs tables and visualizations of u-values against their null references. Together, these outputs provide a structured and interpretable summary of how much model predictions deviate from fairness expectations under a counterfactual group membership assignment.

\subsection{Component 3: Generalized Counterfactual Intersectional Fairness}
The framework in Component 2 was implemented primarily for logistic regression, random forest, and neural network model types, constraining the applicability to a narrow range of architectures and data pipelines. To overcome this limitation, Component 3 introduces a modular estimator structure capable of supporting any Scikit-learn compatible model, including gradient boosting models, tree ensembles, support vector machines, and other custom classifiers. The flexibility afforded by this allows researchers to apply the framework to the model type best suited for their data and greatly improves the accessibility of the framework. \par

Building on this expanded compatibility, Component 3 also extends the original \citet{wastvedt2023} counterfactual intersectional framework to settings without an explicit treatment variable. In the original framework, counterfactual false positive rate (cFPR) and counterfactual false negative rate (cFNR) were estimated by permuting intersectional group labels while holding treatment and other covariates fixed, and in doing so asks: “If all individuals were randomly assigned group membership, how would model error rates vary across intersectional groups, conditional on treatment and covariates?” Under the expansion presented here, group membership is taken as the hypothetical intervention and the framework retains the same counterfactual logic, but without the conditional treatment assignment. The counterfactual error rates are derived similarly from model-imputed probabilities over the observed covariate distribution, and pairwise absolute differences between groups are summarized into aggregate unfairness statistics (average, maximum, variance). These summary statistics are then compared against a group-label permutation derived null distribution representing a hypothetically fair model to yield u-values, quantifying the difference of observed disparities relative to random assignment. \par

Conceptually, this framework enables the quantification of intersectional unfairness even in the absence of explicit treatment interventions. Such an approach is particularly useful in fairness auditing contexts where treatment exposure is not well defined or where interest centers on group-based disparities in model calibration and error rates. However, this framework does inherit several notable conceptual limitations. As group membership itself is typically immutable in real world scenarios, these u-values should be interpreted as model-based fairness diagnostics rather than any sort of causal effects. Additionally, when group membership and the outcome variable are weakly correlated, u-values will trend towards zero indicating model insensitivity to identity as opposed to actual fairness and thus may offer little insight into intersectional disparities inherent in the model itself. In such cases, examining conditional error rates alongside the descriptive disparities metrics offered in Component 1 offers a more reliable basis for interpretation. \par

To help address this sensitivity, the framework offers both single-robust (SR) and doubly-robust (DR) estimation modes. The SR approach relies solely on outcome modeling and is most appropriate when covariates sufficiently capture variance correlated to group membership. The DR approach extends this using augmented inverse probability weighting, similarly to the \citet{wastvedt2023} frameworks utilization of inverse probability weighting, to improve robustness in the case when the outcome or propensity models are mis-specified. The SR approach is generally more stable when group membership has little influence on the outcome, whereas the DR approach improves consistency when at least one model is correctly specified and there exists some level of dependence between group membership and outcome. However, this DR method can also become more sensitive if the group membership and outcome are near independent and may yield uninterpretable u-values or error rates. In practice, applying both methods helps serve as a robustness check where agreement between outputs suggests stable and well-identified estimates, whereas large discrepancies may indicate group membership and outcome independence. \par

\begin{figure}[H]
\centering
\includegraphics[width=\linewidth, height = 0.8\textheight, keepaspectratio]{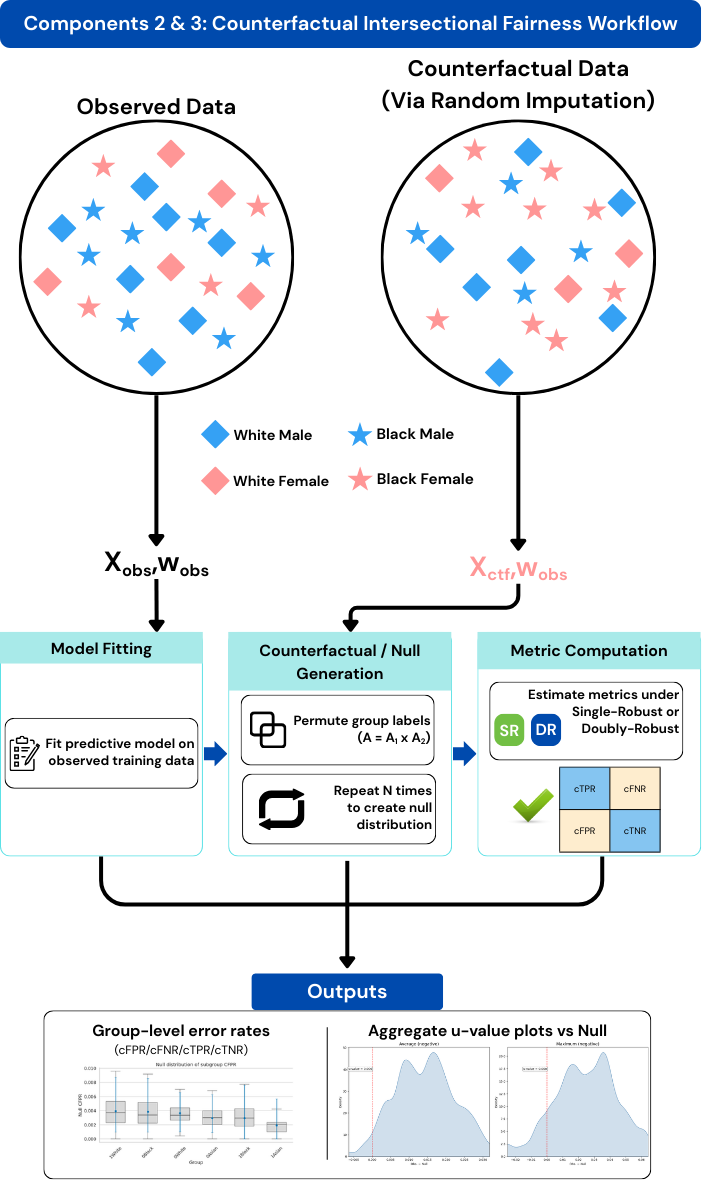}
\caption{\textbf{Components 2 and 3 workflows for counterfactual intersectional fairness analysis.} Observed dataset evaluated using permutation-based null distributions to compute counterfactual error metrics and unfairness (u-values), enabling assessment of whether disparities persist beyond chance.}
\label{counterfactual}
\end{figure}

The general workflow and high-level conceptualizations are similar to Component 2 as described in Figure~\ref{counterfactual}. Differences in the function names and overall class structures are described in detail in the toolkit documentation. 

\subsection{Toolkit Implementation}
The toolkit was implemented using Python version 3.10.11. A full list of the required dependencies, and their versions, can be found in the toolkit documentation and supplemental information. The implementation of the framework was done on both synthetic data, as seen in the package documentation, and real-world clinical data from the \textit{All of Us Research Program} Controlled Tier V8 Dataset. All source code, synthetic data, and documentation for the package is publicly available on GitHub with the repository information found in the supplemental material. All non-synthetic clinical data can be accessed by registered researchers at workbench.researchallofus.org, and scripts detailing the exact data pipeline and application inside the workbench are also available in the GitHub repository. 


\section{Results}

To demonstrate the workflow of the toolkit, the following section will describe the application of the toolkit using real-world clinical data generated in the \textit{All of Us} Researcher Workbench. This application demonstrates how to utilize the toolkit and interpret intersectional disparities in model performance beyond conventional metrics. A more comprehensive description of model features, demographic distributions, applied concepts, and outputs are provided in the supplemental material. \par

The clinical use case is adapted from prior work by \citet{baxter2021} and predicts the need for glaucoma surgery intervention within six months of a participant receiving a glaucoma diagnosis. The original study reported three model types with various hyperparameters, but for this study the best performing logistic regression model that did not indicate significant concerns of overfitting was chosen. The model incorporated 56 predictor variables such as vital signs, physical measurements, and comorbidities available in the EHR data. Protected characteristics were defined as gender and race, yielding multiple intersectional subgroups. Before minority upsampling was applied, the resulting cohort contained 3,880 participants and was predominantly Female (53.2\%) and White (47.1\%). Of this cohort, 726 (18.7\%) had received the outcome (glaucoma surgery) within six months of the diagnosis. Race and Gender were chosen as the intersectional characteristics with our largest populations being White/Male (31.5\%), White/Female (28\%), Black/Female (22.7\%), and Black/Male (17.9\%). A more complete description of the features used, and the cohort characteristics are given in Table~S2 in the supplemental information. \par

The application begins with utilizing Component 1, which provides an observational assessment of model fairness using traditional fairness metrics (demographic parity, equalized odds, equal opportunity difference) in both a single-axis and intersectional manner. This component helps establish a baseline understanding of how model performance varies across intersectional subgroups without imposing any causal assumptions. Application of this component reveals meaningful disparities across intersectional groups. The equalized odds true positive rate gap of 0.33 and false positive rate gap of 0.15 indicate substantial differences in model error rates across intersectional subgroups, suggesting that both sensitivity and false alarm rates vary considerably between groups. The demographic parity gap of 0.20 reflects a twenty-percentage-point difference in positive prediction rates between the most and least favored intersectional groups, indicating notable imbalance in overall model selection rates. In contrast, non-intersectional evaluation reveals smaller disparities. When assessed by race alone, the demographic parity gap decreases to 0.10, with equalized odds gaps of 0.08 for TPR and 0.10 for FPR. Similarly, gender-only analysis shows a demographic parity gap of 0.07, although disparities in TPR remain elevated (0.17), indicating persistent differences in sensitivity. These findings demonstrate that analyses restricted to single protected attributes can mask compounding intersectional inequities that become apparent when demographic categories are jointly evaluated. \par

Taken together, these results suggest that while overall predictive performance remains moderate (AUROC = 0.709, accuracy = 0.651), the model’s benefits and errors are not evenly distributed across intersectional subgroups. A sample of the output metrics is shown in Table~\ref{tab:tprfpr} and Table~\ref{tab:fairnessgaps}, with complete intersectional metric tables and inter-group disparity visualizations provided in Table~S3 in the supplemental material.

\begin{table}[H]
\centering
\caption{True positive rates and false positive rates for the non-intersectional and intersectional Race/Gender demographics}
\label{tab:tprfpr}
\begin{tabular}{lcc}
\toprule
Group (n) & TPR & FPR \\
\midrule
Black (300) & 0.69 & 0.40 \\
White (485) & 0.61 & 0.30 \\
Female (433) & 0.72 & 0.33 \\
Male (402) & 0.56 & 0.34 \\
Black | Female (170) & 0.66 & 0.39 \\
Black | Male (130) & 0.72 & 0.43 \\
White | Female (238) & 0.78 & 0.28 \\
White | Male (247) & 0.45 & 0.32 \\
\bottomrule
\end{tabular}

\vspace{0.5em}
\footnotesize{*Results obtained on 20\% held out test set.}\\
\footnotesize{**Demographic groups with $n < 20$ held out.}
\end{table}
\begin{table}[htbp]
\centering
\caption{Model Performance and Fairness Metric Gaps for non-intersectional and intersectional Race/Gender demographics}
\label{tab:fairnessgaps}

\begin{tabularx}{\linewidth}{lXXX}
\toprule
Metric & Race Only & Gender Only & Intersectional \\
\midrule
Demographic parity gap & 0.1038 & 0.0692 & 0.200 \\
Equalized odds FPR gap & 0.1031 & 0.0133 & 0.1462 \\
Equal opportunity gap & 0.0796 & 0.1661 & 0.3316 \\
\bottomrule
\end{tabularx}

\vspace{0.5em}
\footnotesize{*Model Accuracy: 0.651}\\
\footnotesize{**Model AUROC: 0.709}
\end{table}

\begin{figure}[htbp]
\centering
\includegraphics[width=0.9\linewidth]{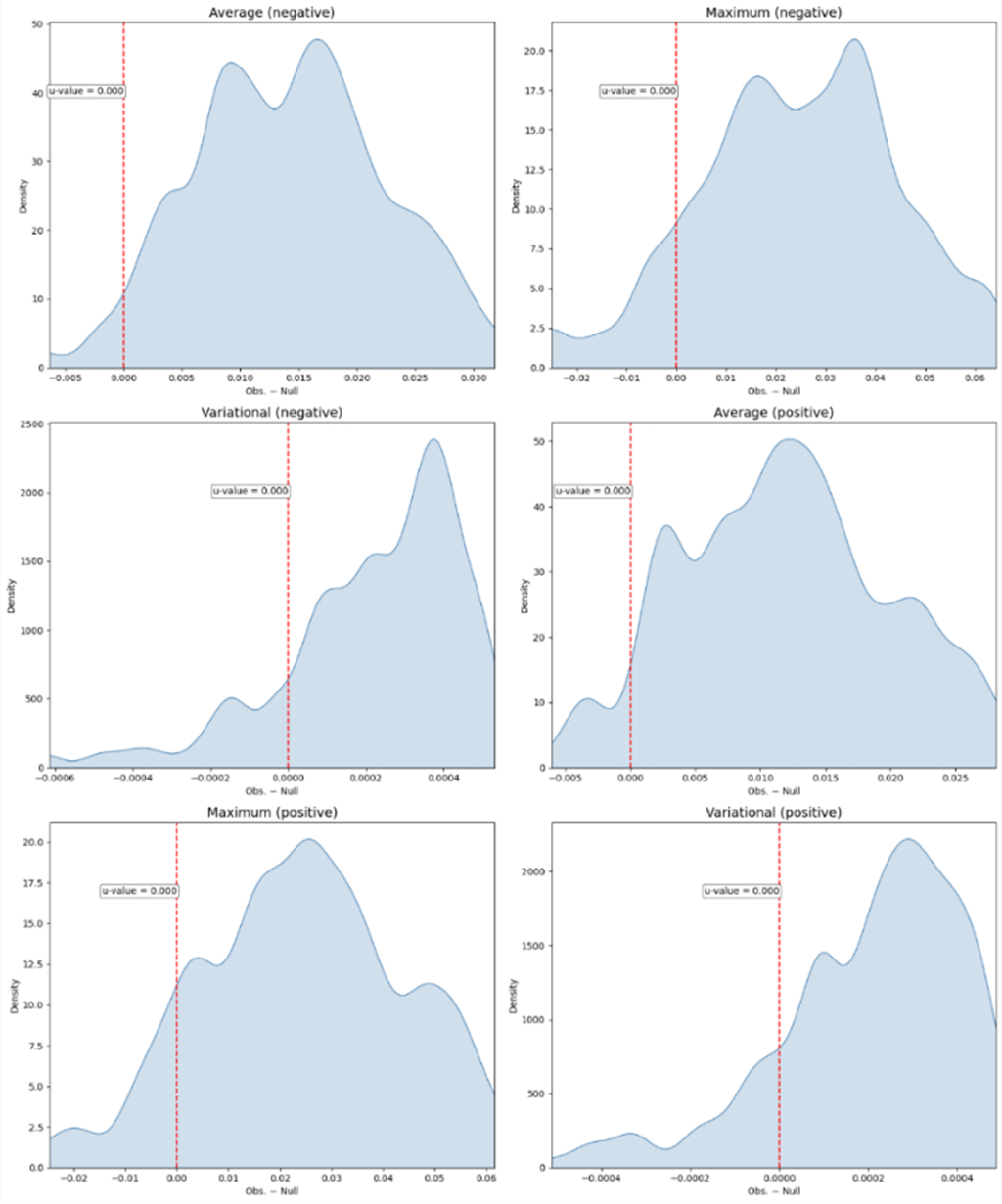}
\caption{Plots from Component 3 showing the aggregate disparity metrics with the u-values overlaid in red. Threshold for acceptable unfairness set to 0.1.}
\label{component3_uvalues}
\end{figure}

Although Component 1 provides invaluable insight into disparities in the model, these results are observational and cannot conclusively determine whether the observed gaps in model performance arise from spurious correlations or from systemic bias embedded in the model itself. To further evaluate these model performance disparities, the next step is application of Component 3 to assess how much of the unfairness is attributable to group membership. For this next component, the model and data stay the same and the SR estimation method is used with bootstrap resampling enabled and 200 iterations of the permutation-based null distribution generation. This next step holds the fitted model fixed and simulates counterfactual group assignments through permutation that is compared with the observed model to generate u-values quantifying the degree to which observed disparities exceed what might occur by chance under the null hypothesis. The output of this component contains several plots showing the difference between the observed outcomes and the null distribution, with the u-value for each metric overlaid (average, max, variance) as well as intersectional group-level plots showing the cFPR/cFNR. These u-value plots are shown in Figure~\ref{component3_uvalues}.

Results from Component 3 indicate that the intersectional disparities identified in Component 1 cannot be conclusively attributed to systematic dependence of model predictions on group membership. Using a prespecified fairness threshold of 0.1, the estimated u-values trend towards 0 across all metrics suggesting that the observed disparities were statistically indistinguishable from those expected under the permutation-based null distribution. In other words, conditional on covariates, all groups have the same likelihood to be correctly predicted, even when accounting for random variation in group assignment.

\section{Discussion}

The findings from this study highlight several important contributions towards advancing practical approaches to intersectional fairness evaluation in clinical machine learning workflows. The \textit{Fairlogue} toolkit demonstrates that intersectional fairness auditing can be implemented within a modular analytical framework that integrates both observational and counterfactual evaluations. This design enables practitioners to extend beyond traditional single-axis evaluations and systematically examine how disparities arise through complex interactions between demographic characteristics. In operationalizing these analyses within a single unified workflow, the toolkit supports more accessible, and reproducible fairness assessment in applied clinical contexts. \par

The glaucoma surgery prediction task results illustrate the practical value of the toolkit. Intersectional analysis revealed disparities that were substantially larger than those detected through single-axis evaluations, reinforcing longstanding concerns that traditional fairness auditing may underestimate inequities affecting marginalized populations. Despite moderate predictive performance (AUROC = 0.709; accuracy = 0.651), the distribution of errors and positive predictions varied meaningfully across race-gender intersections. These findings underscore a central insight of intersectionality: fairness assessments that isolate demographic attributes risk obscuring compounded disadvantages that only become visible when jointly examined. From a translational perspective, such masked disparities could lead to unequal access to downstream interventions, differential clinical attention, or unintended reinforcement of structural inequities. \par

The counterfactual components of the toolkit serve to provide an additional layer of diagnostic rigor. Comparing observed error rate disparities to rates observed under a counterfactual, hypothetically perfectly fair scenario allows practitioners to assess whether disparities persist beyond what would be expected under random group assignment. In the present case study, counterfactual unfairness estimates approach zero under specified thresholds, suggesting that conditional on covariates, the model’s predictions did not exhibit systematic dependence on intersectional group membership. This evaluation illustrates the importance of distinguishing between observed disparities and disparities attributable to model-based unfairness. Such distinctions are essential for informing appropriate mitigation strategies, as interventions may differ depending on whether inequities arise from model structure, data imbalance, or even broader structural processes embedded in the data or pipeline. \par

However, interpretations of counterfactual fairness metrics require caution. When protected characteristics are only weakly associated with the outcome, these frameworks may lack sufficient signal to detect systematic bias, resulting in $u$-values that converge towards zero regardless of model structure. Under these conditions, low unfairness estimates should not be interpreted as definitive evidence of fairness but rather as reflecting limited identifiability of group-based dependence within the data. In such cases, observed disparities may arise not from direct effects of intersectional group membership alone, but from more complex relationships between other confounding variables and these demographic characteristics. This highlights a broader methodological consideration: counterfactual fairness diagnostics should be interpreted alongside descriptive disparity metrics. In practice, the combined use of observational and counterfactual components provides a more robust fairness evaluation than either approach alone. \par

Accordingly, several limitations should be considered. First, reliable fairness auditing depends on adequate subgroup representation. This remains a concern in both single-axis and intersectional evaluations but is particularly relevant when the dimensionality of protected attributes increases as subgroup sparsity will limit statistical precision and stability of error rate estimates. Although \textit{Fairlogue} includes mechanisms to mitigate instability, practitioners should remain attentive to sample size constraints when interpreting results. Second, as previously mentioned, counterfactual estimation assumes that observed covariates sufficiently capture variation associated between group membership and outcome risk. Significant confounding influence by other variables or model misspecification may therefore influence fairness diagnostics despite the inclusion of single-robust or doubly-robust estimation methods. Third, like all auditing frameworks, \textit{Fairlogue} only evaluates fairness conditional on an already trained or defined model. Other than a Youden’s based prediction threshold option made available in Component 3 for convenience, it does not inherently implement any bias mitigation techniques. \par

Despite these considerations, the broad utility of the toolkit lies in how it operationalizes intersectional fairness auditing for both experts and non-specialists. The technical burden of implementing both pre- and post-processing techniques for intersectional subgroups can be substantial; a burden that is alleviated by the modular workflow presented here. Consolidating descriptive analytics (Component 1), Counterfactual Group Membership in Causal-Treatment analysis (Component 2), and Counterfactual Group Membership analysis (Component 3) within a single, unified package enables users to progress from diagnosing disparities to testing their counterfactual persistence with minimal expertise required. This accessibility has significant implications for responsible AI/ML development by empowering a wide audience of users to rigorously evaluate model fairness in an intersectional manner within their respective domains. \par

While this toolkit has been demonstrated primarily on electronic health records and synthetic clinical datasets, the methods are generalizable to any other field where intersectional inequalities may be prevalent, including finance, education, and criminal justice. Comprehensive documentation, example use cases, and reproducible synthetic datasets are included in the package to help users understand, implement, and even extend the toolkit to their own needs and development environments.

\section{Conclusion}
The\textit{ Fairlogue} toolkit represents a significant development in accessibility of the algorithmic fairness field by providing tools for any practitioner, model developer, researcher, or other stakeholder to quickly and easily assess whether their machine learning model is perpetuating biases. In doing so, this tool will not only ensure that the models that are being developed and deployed are equally fair for not just single-axis protected characteristics but for all intersectional demographics. This tool is made publicly available, and accessible to users of all skill levels, to increase the visibility and reach of the entire algorithmic fairness field. The impacts of this toolkit extend not just to health informatics, but in all fields where machine learning models are implemented on data with protected or vulnerable populations. Looking ahead, by virtue of being open sourced and commitment to community participation, \textit{Fairlogue} is well situated for continued development and expansion as the research interest into intersectional approaches continues to grow.

\section{Acknowledgements}
We gratefully acknowledge \textit{All of Us} participants for their contributions, without whom this research would not have been possible. We also thank the National Institute of Health’s \textit{All of Us Research Program} for making available the participant data examined in this study.


\end{document}